# Expressing Facial Structure and Appearance Information in Frequency Domain for Face Recognition

C. C. Olisah, S. Nunoo, P. Ogedebe and Ghazali Sulong

**Abstract** Beneath the uncertain primitive visual features of face images are the primitive intrinsic structural patterns (PISP) essential for characterizing a sample face discriminative attributes. It is on this basis that this paper presents a simple yet effective facial descriptor formed from derivatives of Gaussian and Gabor Wavelets. The new descriptor is coined local edge gradient Gabor magnitude (LEGGM) pattern. LEGGM first uncovers the PISP locked in every pixel through determining the pixel gradient in relation to its neighbors using the Derivatives of Gaussians. Then, the resulting output is embedded into the global appearance of the face which are further processed using Gabor wavelets in order to express its frequency characteristics. Additionally, we adopted various subspace models for dimensionality reduction in order to ascertain the best fit model for reporting a more effective representation of the LEGGM patterns. The proposed descriptor-based face recognition method is evaluated on three databases: Plastic surgery, LFW, and GT face databases. Through experiments, using a base classifier, the efficacy of the proposed method is demonstrated, especially in the case of plastic surgery database. The heterogeneous database, which we created to typify real-world scenario, show that the proposed method is to an extent insensitive to image formation factors with impressive recognition performances.

**Keywords** Facial structure; Appearance information; Frequency domain; Face representation; Face recognition. Plastic surgery

## 1 Introduction

Pattern representation is still an important problem in the field of computer vision, machine learning and psychophysics. A number of real-world applications, e.g., pattern classification and recognition, video and image retrieval, object segmentation and detection and scene recognition require that salient and discriminative image patterns be represented in the best possible space [1]. To best represent patterns of face images of various classes within a space, they must be distinctive. However, the face image pattern (two dimensional: 2D) unlike other patterns such as fingerprint image, or a natural scene image, has more uncertain primitive visual features in its grey-level state. Even from the pictorial description of fingerprint, natural scene images and their likes, there are usually definite lines, contours, points, edge, texture or shape patterns [2]. It is for this very reason that pattern representation is still an important topic in face recognition.

The need to effectively identify and represent facial patterns which are deemed discriminative enough for recognition are more heightened when any of the factors such as illumination, pose and scale (these are referred to as the scene-centric conditions); expression, disguise and make-up (these are referred to as the reversible appearance-centric conditions) aging and plastic surgery (these are referred to as the non-reversible appearance-centric conditions) are present. We take for instance plastic surgery; a case of the non-reversible appearance-centric conditions. Would it not be interesting to be able to obtain facial patterns that describes a person's facial patterns which do not change with plastic surgery? so as to obtain features that can be used to recognize a person even after undergoing plastic surgery. For this to be feasible, a pattern representation approach should be able to 1) exploit the actual face image pattern, which we coined as the primitive inherent structural patterns (PISP) of the face, 2) be able to express the facial PISP and global cues discriminative characteristics in the frequency domain, and most importantly 3) be able to retain them in the reduced space in order to effectively recognize the face image of a class from those of other classes.

With the common goal to tackle the problems in face recognition, different and yet related research disciplines, have emerged with numerous face recognition methods. However, these methods are limited by their ability to only address partly the previously emphasized idea of what a good face representation should encompass. For ease of comparison, we restrict our discussion to face recognition methods under the following categories: holistic based representations and local appearance based representations. The earliest holistic based representations stem from the subspace methods, such as PCA [3] and LDA [4]. PCA projects the principal components, usually described as the eigenvectors, linearly along the direction of maximal variance [3]. PCA, as feature extraction method [5], results in poor description of face image features since the data variance might tend towards the direction that is not related to facial patterns and appearance. With the inability to exploit the nonlinear nature of relevant data structures, PCA suffers from sensitivity to a number of facial variation factors resulting from scene-centric conditions or/and appearance-

Manuscript submitted on April 28, 2017.

C. C. Olisah was with Virtual, Visualization and Computer Vision (Vicube) Research Lab, Software Engineering Department, Universiti Teknologi Malaysia. She is now with the Department of Computing, Baze University, Nigeria (e-mail: chollette.olisah@bazeuniversity.edu.ng).

S. Nunoo is with the Dept. of Electrical and Electronic Eng., University of Mines and Technology, Ghana (e-mail: sonunoo@umat.edu.gh)

P. Ogedebe is with the Department of Computing, Baze University, Nigeria. (e-mail: pogedebe@bazeuniversity.edu.ng).

G. Sulong is with Virtual, Visualization and Computer Vision (Vicube) Research Lab, Software Engineering Department, Universiti Teknologi Malaysia. (e-mail: gsulong@utmspace.edu)

centric conditions. The limitations of PCA are likewise shared with all holistic based representation methods. However, the LDA is able to find application to classification tasks [4] due to its ability to maximize the separability criterion of between-class scatter in relation to the within-class scatter. The later years saw the introduction of the non-linear methods; isometric mapping (ISOMAP) [6], discriminative information preservation (DIP) [7], maximum variance unfolding (MVU) [8], Laplacian eigenmaps (LE) [9] and its linear counterpart, the locality preserving projection (LPP) [10], its supervised version (sLPP) [11], and locality sensitive discriminant analysis (LSDA) [12]. A general framework, namely linear graph embedding (LGE) and orthogonal linear graph embedding (OLGE) has been proposed as a platform within which some of these methods can be integrated [13]. The advantage of the non-linear methods over the linear methods is that they are able to handle non-linearly spaced data, though this comes at a price. When a subspace from sets of data is composed of discontinuity in data distribution, the non-linear methods often fail [14]. The non-linear methods have been observed in a number of literatures, e.g., [14, 15] to fail to outperform their linear counterparts for such a complex scenario. Since the holistic based representation methods are generally known to perform poorly as feature extraction methods, they are mostly applied as dimensionality reduction methods.

Under the local appearance based representations, different levels of information that are not usually apparent in grey-level (intensity description) face images are extracted so as to emphasize on local details, e.g., texture and/or shape information. However, the type of the local details retained plays a vital role in face recognition tasks, especially in complex instances where many appearance variation factors are entangled. Though humans have demonstrated the capability to ascertain identity in such a complex scenario, computers are yet to operate at such a high degree of face detail analysis. The efficacies of the local appearance based representation methods have been demonstrated in a number of literatures of which the LBP and Gabor are proven to be the most successful and highly applied in face recognition task. LBP describes a central point pixel by the changes in its neighboring pixels. One of the earliest works that used LBP for face representation and recognition is that of [16]. Their work demonstrates that the hidden texture components in the form of micro-patterns of the face image are beneficial to face recognition. To that effect, many LBP-based approaches have been proposed in the literature. Some of which are the local binary pattern histogram Fourier features (LBP-HF) [17], completed local binary pattern (CLBP), which comprises of CLBP-M-S (magnitude and phase), CLBP-S (phase), CLBP-M (magnitude) [18] and LBP in pyramid transform domain (PLBP) [19]. Describing facial appearance by accumulating pixel level neighborhood relationships on a holistic level or block level preserves image local details. These local details are termed micro-textons or micro-patterns by [16] and are known to be well established technique for texture description. This approach is the principle of LBP and it is extended to its variants. Even though a number of good performances have been reported in literature for well-researched databases, LBP and its variants have their weaknesses. Most importantly, the pixel level neighborhood relationships enhance on image texture properties and so might fall short where non-reversible image formation factors like plastic surgery are present [20].

The Gabor descriptor and its variants; Gabor is known to mimic human visual cortex and so it is able to encode facial shape, appearance, and texture (but on a coarser scale than LBP [21]). The salient visual properties, such as spatial localization, orientation, selectivity, and spatial frequency characteristics of Gabor wavelets, make it a powerful tool for face pattern representation and recognition. In recent times, many variants of Gabor have been proposed to improve performance and handle specific difficulties in face recognition. Some of the Gabor based descriptors are; histograms of Gabor ordinal measures (HOGOM) [22], local Gabor binary pattern histogram sequence (LGBP) [23], local Gabor XOR patterns (LGXP) [24]. Beyond doubt, Gabor and its variants have shown their validity in tackling specific difficulties related to scene-centric problems such as illumination, occlusion [25], and pose [26], and appearance centric problems such as facial expression [11, 27, 28], face mutilation [29], and plastic surgery [20, 30]. It should be noted that Gabor and its variants, acting on grey-level (intensity) image [31], emphasize on texture properties of the image as opposed to shape. From the studies [32-33] magnitude component of Gabor wavelets expresses discriminative information.

Our emphasis from these reviews is that in the case where the image formation factors such as the ones arising from scene-centric conditions and reversible conditions are the only factors considered then careful thought into the representation approach of face image pattern should not be of much significance because they can be controlled. However, it renders the face recognition system impractical because instances of non-reversible conditions, which cannot be controlled, might form a big part of the system. With this problem, the descriptors that emphasize on structure-less facial details as features will fall short. Therefore, to be able to address this problem we take into consideration the four solution strategies in the design of an effective representation approach.

Our contributions: We hypothesize that the primitive inherent structural pattern (PISP) of the face which can be extracted using the derivative of Gaussian is an essential cue in face description and makes noteworthy recognition improvement to Gabor. We propose a simple yet efficient face descriptor approach coined local edge gradient Gabor magnitude (LEGGM) pattern so as to express the PISP and global appearance of face information in frequency domain using the Gabor wavelets. We adopt linear subspace models for dimensionality reduction and analyze experimentally which model strategy is able to retain the descriptor data in the reduced space, that is, the best fit model for effective representation and further recognition.

The rest of the paper is organized as follows. In section 2, the proposed LEGGM descriptor and the art of describing a person's unique facial pattern using LEGGM is presented. In Section 3 we briefly introduce subspace models for the purpose of reducing the dimensionality of the descriptor data. In section 4, the experimental application scenario is presented while in section 5 is the experimentation results and discussion. In section 6 is the conclusion.

## 2 Face Description

The algorithmic process for extracting the local edge gradient Gabor magnitude (LEGGM) pattern is illustrated in Figure 1. Given the illumination normalized face image, the actual processing for LEGGM descriptor comprises of the following: a) determining PISP, b) Complete face structural pattern computation, c) Expressing output of (b) in frequency domain, d) Down-sampling, and E) Normalization.

### 2.1 Determining the Primitive Inherent Structural Pattern

In the following discussions, the PISP computation through derivative of Gaussian processes will be mathematically formulated.

Having defined a set of two-dimensional (2D) Gaussian filters $7 \times 7$ and $13 \times 13$. We go on ahead to derive them based on convolution, using a $3 \times 3$ Laplacian kernel. However, since the objective is to convolve over discrete sample points of a sample $I'(c)$, the following derivation is established:

$$f(c) = G(c;\sigma) * I'(c) \qquad (1)$$

where $G(c;\sigma) = \frac{1}{2\Pi\sigma^2 r} e^{-\frac{\|c\|^2}{2\sigma^2 r}}$ is a Gaussian kernel, and $\sigma = (\sigma_1, \sigma_2)$, which is the standard deviation. The symbol '$*$' denotes a convolution operator, the norm of $c$ is given as $\|c\| = \sqrt{x^2 + y^2}$. The Gaussian kernels $G(c;\sigma_1)$ and $G(c;\sigma_2)$. On applying the Laplacian kernel [34] given as $\nabla^2$, (1) becomes [33]:

$$\nabla^2 f(c) = \nabla^2 (G(c;\sigma) * I'(c)) \qquad (2)$$

where $\nabla^2$ denotes that derivation is the second order operation. It should be noted that we adopted the negative Laplacian kernel. Since convolution is associative, which means that whether the image $I'(c)$ is convolved with the Gaussian kernel $G(c;\sigma)$ before differentiation or the Gaussian kernel is differentiated before convolution with $I'(c)$, the same output will always result. By this associative property, (2) can be re-written as:

$$\nabla^2 f(c) = (\nabla^2 * G(c;\sigma)) * I'(c) \qquad (3)$$

Typically, convolution is a superimposition operation across the pixel coordinates of the image. Hence, the entire operation can be expressed as:

$$\nabla^2 f(c) = \sum_i (\nabla^2 G(c-i;\sigma)) I'(c-i) \qquad (4)$$

where $(c-i)$ denotes that weighting is by a shift operation across pixel coordinate $c$ $(x, y)$ upon which the kernel is superimposed. Hence, with respect varying sizes of Gaussian $\sigma_1$ and $\sigma_2$, we represent the resulting outputs in terms of $x$ and $y$, Equation (4) can be re-written as:

$$\nabla^2_x f(c) \approx \frac{\partial^2}{\partial x^2} f(c) = \sum_i \frac{\partial^2}{\partial x^2} G(c-i;\sigma_1) I'(c-i) \qquad (5)$$

$$\nabla^2_y f(c) \approx \frac{\partial^2}{\partial y^2} f(c) = \sum_i \frac{\partial^2}{\partial y^2} G(c-i;\sigma_2) I'(c-i) \qquad (6)$$

Note that, since $\sigma$ controls the spread of the Gaussian distribution and its resulting smoothing effect on an image, $\sigma_2$ must be greater than $\sigma_1$, but related to $\sigma_1$ in order to efficiently capture the actual gradient revealing the PISP, which is preserved by the difference of the derivative of Gaussians.

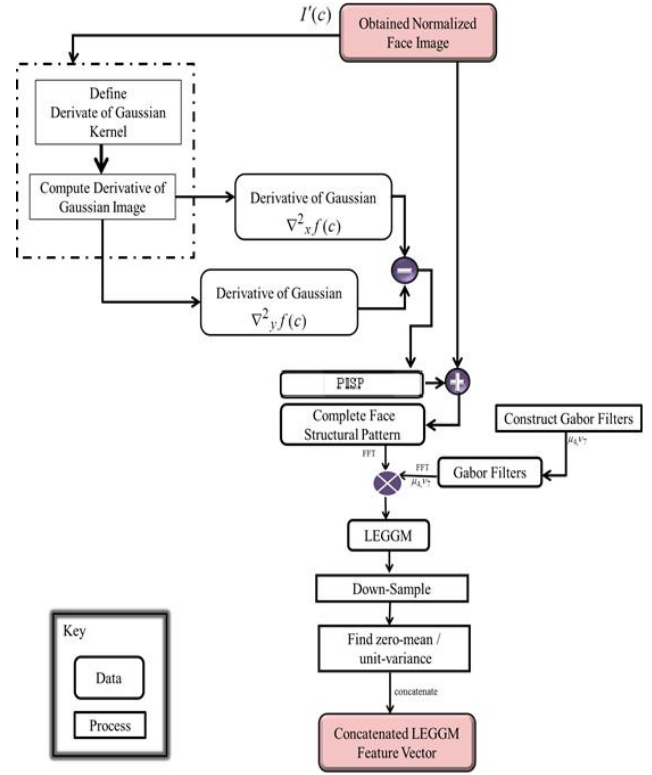

**Figure 1** The descriptor algorithmic process

More also, it should be noted that instead of padding with zeros, the kernels at the borders are replicated to fill the points outside the kernel. The derivative kernel plays a major role in the resulting derivative of Gaussian kernel since its major role is to enforce zero-crossing effect of the Laplacian, where only rapid change is highlighted [35]. The resulting derivatives of the Gaussian kernels are important because they both yield different responses on a given image. Precisely, the $7 \times 7$ kernel will respond to smaller details, while the $13 \times 13$ kernel to larger details. However, these different responses are so that on differencing the image filtered using the $7\times7$ derivative of Gaussian filter from the image filtered using the $13\times13$ derivative of Gaussian filter only the actual PISPs are retrieved.

Therefore, the PISP $\Xi$ can be defined as a function of the following:

$$\Xi(c) = \frac{\partial^2}{\partial x^2} f(c) - \frac{\partial^2}{\partial y^2} f(c) \quad (7)$$

Note that based on the filter characteristics of the two derivatives of Gaussians used, which are band-pass filters, (7) translates to the formation of band-pass responses that captures more of coarse details in the image as opposed to fine-details. Having determined the PISP which are not apparent at grey-level, an important step to the descriptor algorithm is to embed the PISP in the grey-level image to create a complete structurally characterized face pattern. Hence, the PISP is accentuated at image grey-level.

### 2.2 Complete Face Structural Pattern Computation

Our intention is to not lose any valuable information. We also capture the global appearance of a sample face by embedding the PISP into the global appearance image, a structurally defined face pattern is created. The embedding process is described as an additive function that calculates at each pixel point of PISP image and global appearance image for the complete face structural pattern. This is mathematically described as follows:

$$\chi(c) = \Xi(c) + I'_\varpi(c) \quad (9)$$

where $\chi(c)$ is the resulting image.

### 2.3 Expressing Patterns in Frequency Domain

We use Gabor wavelet for expressing the frequency characteristics of the obtained patterns from the prior step. Gabor wavelet/kernel has proven useful in pattern representation due to their computational properties and biological relevance [36]. It has become a useful and powerful tool for expressing spatial-frequency information of an image. The Gabor kernels can be expressed as [36]:

$$\psi_{\mu,\nu}(c) = \frac{\|l_{\mu,\nu}\|^2}{\sigma^2} e^{\left(-\|l_{\mu,\nu}\|^2 \|c\|^2 / 2\sigma^2\right)} \left[ e^{i l_{\mu,\nu}{}^c} - e^{-\sigma^2/2} \right] \quad (10)$$

where $\mu$ and $\nu$ define the scale and orientation of the Gabor kernels, respectively, $c = (x, y)$. Here, $\|\bullet\|$ denotes the norm operator, the term $l_{\mu,\nu}$ is the wave vector defined as follows:

$$l_{\mu,\nu} = l_\mu e^{i\phi_\nu} \quad (11)$$

where $l_\mu = l_{\max}/s_f^\mu$ and $\phi_\nu = \pi\nu/8$. $l_{\max}$ is the maximum frequency, and $s_f$ is the spacing factor between kernels in the frequency domain [36] and $\sigma$ is a control parameter for the Gaussian function.

The family of self-similar Gabor kernels $\psi_{\mu,\nu}$ in (10) is generated from a mother wavelet by selecting different center frequencies (scales) and orientations under operations of scaling, translations and rotations [36]. In most cases, the Gabor wavelets at 5-scales ($\mu \in \{0,\ldots,4\}$) and 8-orientations ($\nu \in \{0,\ldots,7\}$) are used. This paper employs these Gabor kernels with the following parameters: $\sigma = \sqrt{2}$, $l_{\max} = 0.25$, $s_f = \sqrt{2}$. The kernels at 5-scales ($\mu \in \{0,\ldots,4\}$) and 8-orientations ($\nu \in \{0,\ldots,7\}$) are generated using these parameters.

The response $O_{\mu,\nu}(c)$ of spatial-frequency Gabor kernels to the complete face structural image is obtained by the convolution of the image with a family of Gabor kernels $\psi_{\mu,\nu}(c)$ at 5-scales $\mu$, and 8-orientations $\nu$. Hence:

$$O_{\mu,\nu}(c) = \chi(c) * \psi_{\mu,\nu}(c) \quad (12)$$

Each $O_{\mu,\nu}(c)$ from (12) is derived via fast Fourier transform (FFT). This implies that [36]:

$$O_{\mu,\nu}(c) = \mathfrak{I}^{-1}\left\{ \mathfrak{I}\{\chi(c)\} \mathfrak{I}\{\psi_{\mu,\nu}(c)\} \right\} \quad (13)$$

where $\mathfrak{I}$ and $\mathfrak{I}^{-1}$ denote the Fourier transform and its inverse, respectively. It should be noted that in the frequency domain, signal details are well preserved and accentuated in a way that is not possible in the spatial domain. Most importantly, signals with strong edges are highly responded to in the frequency domain. In a later part of this subsection, the FFT of an edge defined contained signal ($\mathfrak{I}\{\chi(c)\}$) will be demonstrated alongside the FFT of a plain signal in order to confirm the presented claim.

Typically, by virtue of the FFT, $O_{\mu,\nu}(c)$ is a complex function that is composed of a real part $\xi_{\mu,\nu}(c)$ and an imaginary part $\wp_{\mu,\nu}(c)$. Based on the two parts, the local edge gradient Gabor magnitude pattern can be computed as:

$$LEGGM_{\mu,\nu}(c) = \sqrt{O_\xi{}^2{}_{\mu,\nu}(c) + O_\wp{}^2{}_{\mu,\nu}(c)} \quad (14)$$

Equation (14) shows the absolute values of the frequency information at image location $c$ obtained from summing the squares of $\xi_{\mu,\nu}(c)$ and $\wp_{\mu,\nu}(c)$, which is commonly known as the magnitude responses of the signal.

### 2.4 LEGGM Down-Sampling

Transforming a signal with Gabor is like observing the signal from 40 different perspectives. For the fact that the Gabor transform is a discrete case of the short time Fourier transform, there is the likelihood of the existence of an overlap of signal information [37]. In literature, down-sampling of points across the $128\times128\times40$ Gabor transformed signal, is usually adopted [38] and is also used in this paper for the purpose of benchmarking. However, [37] suggest that the research to remedy the redundancy problem is on-going.

By the two-dimensional bilinear interpolation dependent down-sampling function [39], the $128\times128\times40$ $LEGGM_{\mu,\nu}(c)$ features are scaled to a new size using a down-sampling factor of $p$, where ($p = 64$). The bilinear interpolation when used with a down-sampling factor $p$, considers every 8-by-8 region of $128\times128$ pixels. This is repeated for each of the forty (40) $LEGGM_{\mu,\nu}(c)$ features. Note that on the interpolation of every 8-by-8 region of the $128\times128\times40$ $LEGGM_{\mu,\nu}(c)$ features (image matrix) a $16\times16\times40$ down-sampled $LE\hat{G}GM_{\mu,\nu}(c)$ feature (image matrix) results.

## 2.5 LEGGM Normalization

Given that a single signal is observed from 40 different perspectives and described likewise, it is important to make them tend to a zero-mean and unit-variance. This is a widely used method for Gabor transformed signal data [37] normalization. The zero-mean, unit-variance of the down-sampled data ($LE\hat{G}GM_{\mu,v}(c)$) is formally formulated as [37]:

$$LE\hat{G}GM'_{\mu,v}(c) = \frac{LE\hat{G}GM_{\mu,v}(c) - \overline{LE\hat{G}GM_{\mu,v}(c)}}{\sigma_{LE\hat{G}GM}} \quad (15)$$

where $\sigma_{LE\hat{G}GM}$ is the standard deviation of $LE\hat{G}GM_{\mu,v}(c)$

The very essence of this step is so that the objective functions of the subspace learning models can easily converge linearly to a certain value or range of values for samples of the same class (face images).

## 2.6 Local Edge Gradient Gabor Magnitude Pattern

The local edge gradient Gabor magnitude (LEGGM) pattern at pixel position $c$ for an $i$th image sample is formally defined as follows:

$$LE\hat{G}GM'^{i}_{u,v}(c) = [LE\hat{G}GM'^{(T)}_{0,0}(c), LE\hat{G}GM'^{(T)}_{0,1}(c),$$
$$\cdots LE\hat{G}GM'^{(T)}_{4,7}(c)]^T \quad (16)$$

and simplified as,

$$LE\hat{G}GM'^{i}_{u,v}(c) \equiv Z^{i}_{\mu,v} \quad (17)$$

where $Z^{i}_{\mu,v}$ is the augmented features of the forty (40) down-sampled and normalized LEGGM features, which can be used to describe a face image. $T$ is the transpose operator. For simplification, LEGGM algorithmic steps for describing a face sample are pictorially summarized in Figure 2.

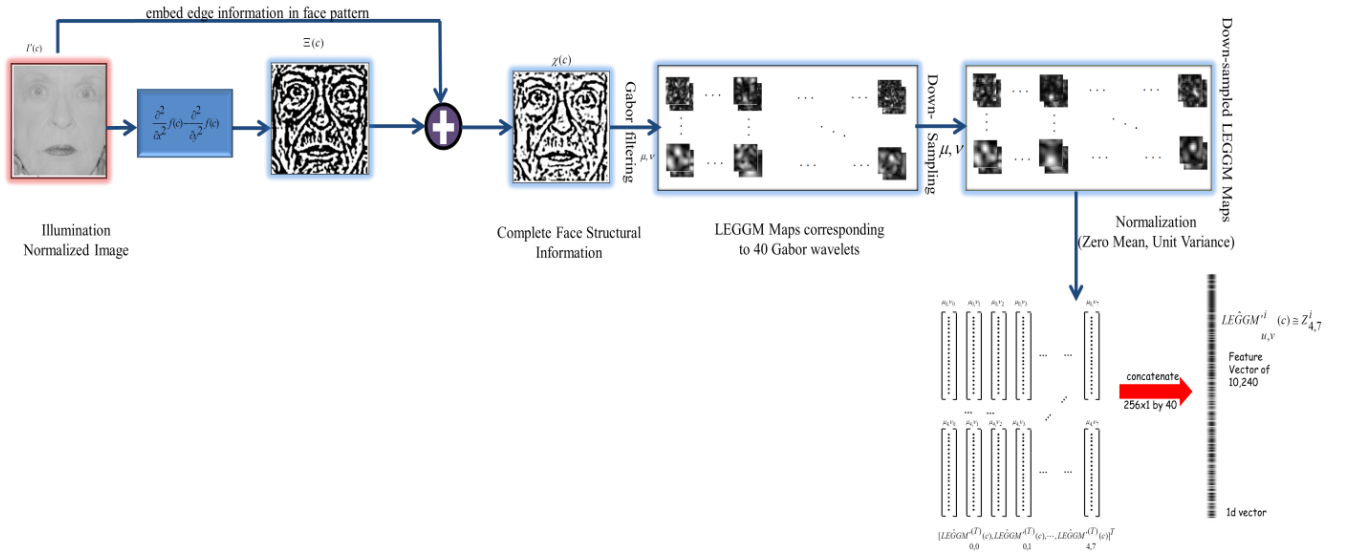

**Figure 2** Pictorial overview of the resulting images of the descriptor main steps

### 2. On Dimensionality Reduction for Best Fit Model

Here we adopt linear subspace methods for dimensionality reduction of LEGGM descriptor data for effective face representation that aids the classification process. We employ the following linear subspace models, principal component analysis plus linear discriminant analysis (PCA plus LDA) [40], locality sensitive discriminant analysis (LSDA) [12], and supervised locality preserving projection (sLPP) [10, 11] independently for learning LEGGM data. These subspace learning models are well used in literature and they operate on various mode of finding their objective functions. LSDA and sLPP are both described under a general framework known as LGE and OLGE [13]. First of all, a brief introduction of the common goal of these models is presented. Then their respective objective functions for arriving at the said goal are discussed.

Given the high-dimensional LEGGM data, let $Z \in \Re^D$ define a set of training samples from a sample set of class $c_i$, where $D$ denotes the high-dimensionality of their vector space. Under the assumption that each class is of unknown distribution, the optimal interest of the linear subspace models is to obtain a projection matrix $W^T$. This matrix results from the mapping of the original high-dimensional data of the feature vector in $D$-dimensional space onto a $d$-dimensional space by a transformation model $\kappa$ expressed as:

$$\kappa : \Re^D \to \Re^d \quad (18)$$

where $\kappa$ can be any of the subspace learning models (PCA plus LDA, LSDA or sLPP). Usually $d$ is of much lower dimension than $D$, that is, $d \ll D$. The significance of the projection matrix $W^T$ is to project a new sample data (a probe or test sample) that is an input to the face recognition system. The new feature vector $y \in \Re^d$ of a sample image described by LEGGM is obtained:

$$y = W^T A \tag{19}$$

where $A$ is the projected sample.

Therefore, the optimal objective of the subspace learning models is to search $W^T$, a learned matrix representation in which all the *significant* observations are well retained. However, the process of finding such a matrix varies with different subspace learning models.

## 3. Experimental application scenario

Face recognition task for application purposes can be defined as a function of face identification and verification. While some of the application areas can be strictly categorized under identification task or verification task, some of them cut across the two tasks. The airport application scenario is one such example. To effectively represent a practical scenario like the airport, various color image data sets are used and this will be discussed in succeeding subsections.

### 4.1 Plastic Surgery Data Set

The plastic surgery data set [41] contains near frontal faces of real people who have undergone plastic surgery. In all, there are a total of 1800 face images of 900 subjects (excluding cheek and chin surgery procedures with 21 subjects, i.e., 42 samples). We created a mirror reflect of the 921 subjects face images. Two experimental scenarios (ES) are defined as follows. ES-1: Three images per subject, one each for the train set, gallery set and test set (the test set is unseen during the training phase). ES-2: Four images per subject, three images per subject are used to make up the train set and also used for the gallery set. The remaining image unseen during the training phase is used to make up the test set.

### 4.2 The Georgia Tech Face Data Set

The Georgia Tech face data set [42] contains about 750 face images of 50 subjects. The experimental scenario is given as follows. ES-1: Out of the 15 face images per subject, 14 face images are selected for train set/gallery set and a single face image that is unknown to both the train/gallery sets is used to make up the test set.

### 4.3 Labelled Faces in the Wild Data Set

The LFW data set [43] comprises of 13233 color face images of 5749 people, out of which a total of 873 people have at least three images, 610 people have at least 4 images and 158 people have at least 10 images. In ES-1: Out of the 158 subjects with at least 10 images, 7 images are selected for train set, while 3 images selected arbitrary to make-up the gallery set (note there is overlap between train and gallery set) and the remaining single image that is unknown to both the train and gallery sets is used to make up the test set. In ES-2: Again out of the 4 images per 610 subjects, 3 images are selected for train set, while the remaining 1 image is used to make up the test set.

### 4.4 Heterogeneous Data Set

In this data set, images of different subjects from the plastic surgery data set are selected arbitrarily from the different subsets, which include every plastic surgery procedure, making a total of 321 subjects with plastic surgery cases. Then full frontal faces with illumination problem of 231 subjects are selected from the Essex data set. An additional 50 subjects are added from the GT data set, and 38 subjects from the LFW data set. This brings the total number of subjects to 640, with every subject having 3 images. ES-1 is given as: 2 images are used to make up the train/gallery set, while the remaining image makes up the test set (probe). For all the subjects the image selected for the test set is unseen during the training phase.

## 4. Experimental results

Firstly, experiments are run starting with the plastic surgery database, followed by GT, LFW and the heterogeneous database. In all the experiments the identification results and verification results are reported using the cumulative match characteristics (CMC) curve, receiver operating characteristics (ROC) curve or points from the ROC curve, and the equal error rate (EER) evaluation metrics.

### 5.1 Evaluation and Benchmarking of LEGGM with Contemporary Face Descriptors

Using ES-2 of the plastic surgery database, the identification results of different descriptor-based face recognition methods are presented without employing any subspace learning/training. The descriptors are used in their original feature-dimension. The facial descriptors under comparison are the LBP variants, which are the CLBP-M-S, CLBP-M and CLBP-S, while the Gabor variants used are LGBP and LEGGM. The identification rates are reported on Rank basis, where the Ranks 1-10 are considered. The results of employing different facial descriptors in the recognition of faces that have undergone plastic surgery are reported for various plastic surgery procedures and their results shown in Figure 3. From the figure the following observations are made.

The Gabor based descriptors are observed to be more robust against non-reversible facial appearance changes due to plastic surgery procedures. Their robustness is shown by their above 65% Rank-1 recognition rate that they achieved in a number of the experiments, which is more than what the LBP based descriptors achieved. The identification accuracy of LBP based descriptors are rather disappointing. They failed to reach a satisfactory recognition rate despite existing in a much lower-dimensional space. Overall, LEGGM, a facial shape and appearance descriptor, shows to have achieved the best Rank-1 identification rates. Its highest Rank-1 identification rate is above 87%, which is achieved for the case of recognizing faces that have undergone Dermabrasion surgery.

While surgery procedures to some facial features such as the eye, nose, forehead and the entire-face (which have been found in psychophysics and computer vision, to contribute largely to face recognition accuracy [44]) minimally affects outlines of the facial features. More of the effects are to the skin regions surrounding the features where the stretching of skin is done to achieve aesthetics. For surgeries that involved such procedures only a minimum-maximum of 8% and 76% correct identification rates were observed for all the descriptors compared. Though, the best performing descriptor is LEGGM facial shape and appearance descriptor, its Rank-1 identification capability did not go beyond 76% for the cases of Blepharoplasty (eye), Rhytidectomy (entire-face), brow-lift (forehead and eye) and Rhinoplasty.

Observed also in Figure 3 is that LEGGM is mostly unaffected by skin texture changing plastic surgery procedures. The identification rates for texture changing procedures reached 87.50%. The closeness in performance of LGBP to LEGGM shows that they share something in common in comparison with the CLBP-M-S [19], CLBP-M or CLBP-S [19]. The CLBP-S performed surprisingly well from Rank 5 to 10 in the recognition of faces that have undergone Blepharoplasty surgery, while LGBP [23] performed the best from Rank-2 to Rank-10 in the recognition of faces that underwent cheek and chin surgery. Both identification performances of LEGGM and LGBP for the cheek and chin surgery altered faces may not be unconnected with their performances achieved for the texture changing procedures because the region that is modified after chin surgery is not included in the cropped face image.

From Table 1, LGBP, CLBP-M-S, CLBP-M and CLBP-S show that they are most appropriate for face verification task than recognition task. Their performances in verification task differ greatly from their performances in the identification task. For instance, take the case of Rhytidectomy where the CLBP-M achieved as low as 8.44% identification rate. In the verification task it achieved as high as 84.09%, 52.60%, 69.81% and 76.62%, verification rates at points on the ROC curve where FAR is 0.1591 (EER), 0.01, 0.05 and 0.1, respectively. Similar performances are observed for the other descriptors such as LGBP, CLBP-S, and CLBP-M-S.

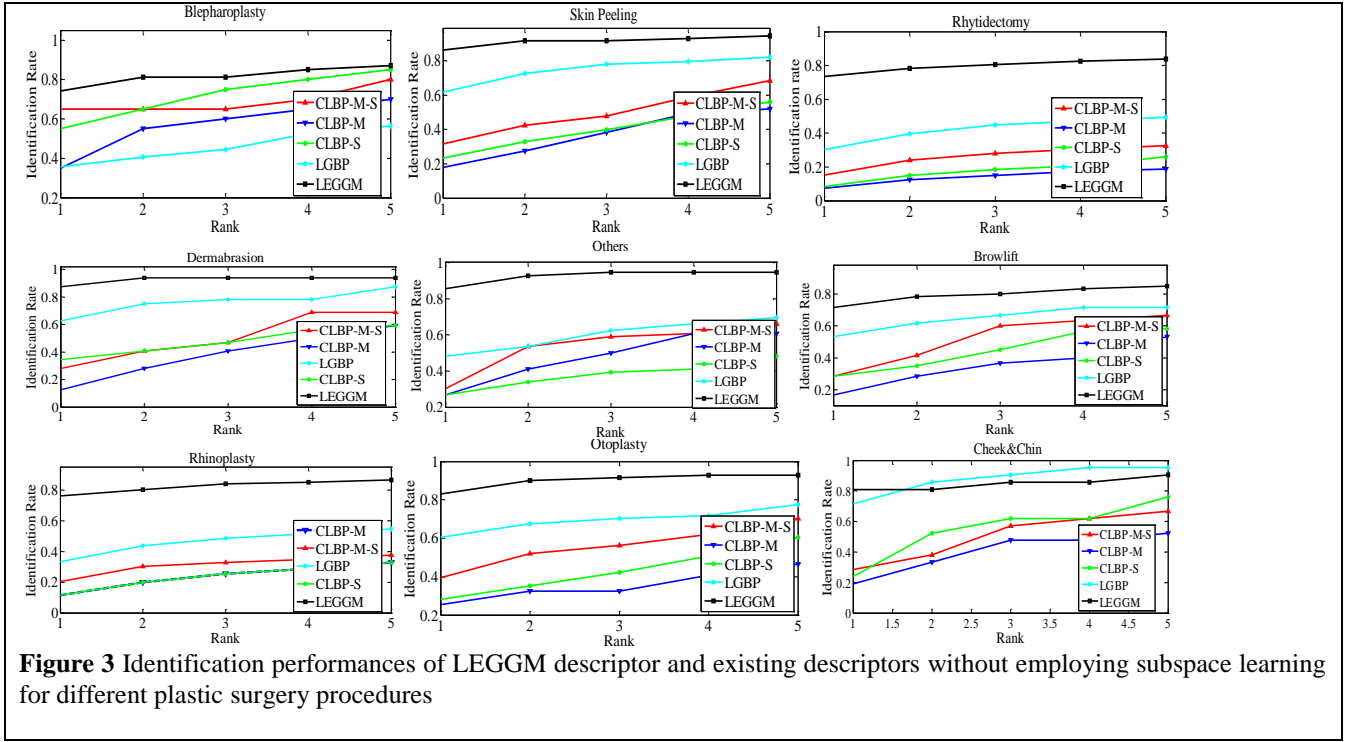

**Figure 3** Identification performances of LEGGM descriptor and existing descriptors without employing subspace learning for different plastic surgery procedures

## 5.2 Experimental Analysis and Performance Evaluation of LEGGM with Linear Subspace Methods

Having ascertained the superiority of LEGGM over contemporary descriptors without training samples, we go ahead to report its performance on employing linear subspace methods. Of interest in this experiment is to report the performances of LEGGM with dimensionality reduction. LEGGM feature vector lies in high-dimensional space and as such hinders effective classification and recognition. However, one cannot sufficiently justify without experimental analysis the subspace models that best fits LEGGM data. On this basis, systematic analysis of the identification and verification performances of LEGGM on employing different subspace learning methods are presented. The subspace learning methods employed are the linear subspace methods which are PCA plus LDA [40] [47], sLPP [11] and LSDA [12]. The performances of sLPP and LSDA are obtained under a generalized LGE and OLGE framework [13]. ES-1 and ES-2 are used here to obtain the verification and identification results. The identification rates obtained for LEGGM using the linear subspace models are plotted individually for all the plastic surgery procedures from Rank 1-5. Since many computations were run, the tabulated results of this experiment in Table II only captures the Rank-1 identification rate, verification rate and EER, which are good enough metrics for validating performance. The best rates of the algorithms for each surgery procedure are shown in boldface. The following abbreviations are made, A-ES-2: LEGGM-LPP-LGE, A-ES-1: LEGGM-LPP-LGE, B-ES-2: LEGGM-LPP-OLGE, B-ES-1: LEGGM-LPP-OLGE, C-ES-2: LEGGM-LSDA-LGE, C-ES-1: LEGGM-LSDA-LGE, D-ES-2: LEGGM-LSDA-OLGE, D-ES-1: LEGGM-LSDA-OLGE, E-ES-2: LEGGM-PCA+LDA, E-ES-1: LEGGM-PCA+LDA. From the identification results given in Appendix B (Figure 9) and summarized in Table II alongside the verification results, the following observations are made.

The results of Appendix B (Figure 9) and Table II shows that LSDA perform better in ES-1 than ES-2. Comparing the results of ES-1 with ES-2 suggests that there is a factor that might pose as a hindrance to the performance of LSDA, this factor is the angle difference between the images of a subject. Since the mirror-reflect version image is included in ES-2 and not in ES-1, minimal or no observable increase in the identification rates of LSDA is seen, that is, ES-2 did not improve on the results of ES-1. To a great extent, the verification performances of LSDA in ES-1 is better than its performance in ES-2. For instance, observe the cases of Blepharoplasty, Rhytidectomy, brow lift and Rhinoplasty where the verification rates in ES-1 and ES-2 are as follows: 97.02% and 92.07%, 93.51% and 84.74%, 96.25% and 90.00%, and 94.27% and 91.15%, respectively. The performances of LSDA in both LGE and OLGE frameworks can be explained simply by the fact that LSDA follows a local approach to the globality of LDA by constructing two graphs, a within-class graph and between-class graph, from one nearest neighbor graph [12]. By this property of LSDA it should by all means perform better than PCA plus LDA, but this is not the case, it is rather the reverse. Thus, it is also by no means out of place to say that LSDA is unable to guarantee the global connectedness within the constructed between-class and within-class graphs for data with variability problem. It should be noted that, if globality is considered in constructing these graphs, it can remedy the said limitation of LSDA observed for LEGGM.

Once again, from Fig. 9 and Table II, it can be observed that the simple PCA plus LDA performs far better than LSDA for all the plastic surgery procedures investigated in ES-2. Since PCA plus LDA is concerned with the projection of the within-class and between-class in relation to the global structure of the manifold, it can be claimed that it is somewhat insensitive to the underlying distribution of LEGGM data. In some of the experiments in ES-2 and ES-1, PCA plus LDA performs comparably to sLPP despite its simplicity. Example is the case of skin peeling, brow lift, Otoplasty and cheek-chin in ES-2, while in ES-1 it is the Blepharoplasty, Rhinoplasty and cheek-chin that are observed.

Overall, with reference to ES-1, sLPP under LGE framework outperforms PCA plus LDA and LSDA. SLPP follows a linear approximation of a nonlinear manifold learning method known as the LE [9]. Unlike LSDA, sLPP uses a single graph and class labels to define the local neighborhood information of the data points, which implies that connectedness of data points, can exist. One major advantage of sLPP over other methods compared in this paper is that: it best preserves the essential manifold structure [45] within LEGGM face space, data based on gradient domains have been assumed to follow a Laplacian distribution [46]. This best explains the reason for the best-fit of sLPP (note that this is an observation made from a performance point of view) to the essential manifold structure of LEGGM data. sLPP is a linear approximation of a non-linear dimensionality reduction method, which implies that it is more appropriate for dealing with outliers than PCA plus LDA and LSDA.

### 5.3 Benchmarking of Designed Descriptor-based Method Performance with Contemporary Methods in Plastic Surgery

At this point, it is worth noting that the approaches employed by researchers in plastic surgery face recognition community for the recognition of faces that have undergone plastic surgery vary by their face representation based methods and evaluation methods. However, one common protocol that exists within this community, which stands as the basis for benchmarking, is the fact that the pre-surgery and the post-surgery images are used for evaluation in a train/test setting. On this note, the average of the individual results obtained for various plastic surgery procedures (excluding cheek and chin, because other researchers did not include in analysis) in ES-1 of Table 2 (see Figure 9 Appendix A) will be compared against all the works in the literature that used the whole data set. Summarized in Table 3 are the results of various methods employed by the plastic surgery face recognition research community, which are tabulated alongside the results of the descriptor-based face recognition method(s).

From Table 3 (see Figure 9 Appendix A), it is apparent that the designed descriptor-based face recognition method for the recognition of surgery altered faces performed the best in comparison with all the other methods. The proposed method achieved an increase of 18.07% in comparison to the multimodal method proposed in [47], where a single classifier decision and the entire images in the data set are used. An obvious practice adopted by previous researchers in the recognition of plastic surgery altered faces is the fusion of scores independently obtained from different features or from different regions of interest of a face image. It should be noted that the scores referred here are the classifier- level scores, which means that the match-decisions of a probe sample to the gallery set has been determined using two different features. It is generally known in literature, for example in [48], that the score-level or decision-level feature fusion yields high recognition accuracy than a single feature decision-score. On this basis, it can be stated that the intrinsic facial shape characteristics that LEGGM descriptor captures is experimentally shown to be minimally affected by plastic surgery procedures. The patch-based approach (GPROF) by Liu *et al*. [30], which is also a variant of Gabor, fuses score decisions from multiple feature parts for final classifier decision. In comparison to the designed descriptor-based face recognition method, a difference of 2.59% is achieved, which shows the superiority of the designed method over GPROF despite being of multiple classifier decision. The SSIM (from Table 3) method proposed by Sun *et al*. [49] achieved 77.55% Rank-1 identification rate without employing subspace learning models. In this respect, it can as well be observed from the experimental results in Table 1 that LEGGM (79.83%), which did not include subspace learning, outperforms SSIM by 2.14%.

### 5.4 Performance Evaluation for Other Variation Factors

Under stable conditions, that is, of a frontal view-point (pose) LEGGM descriptor can be said to be robust against other variation factors. However, we did not put limitation to the view-point (pose) of images in the data sets used for the experimental analysis because the change in view-point (pose) is an integral part of the real-world data sets. The results obtained using the databases of GT, LFW and heterogeneous data sets, respectively. The identification rates, verification rates and EER results are presented here.

#### 5.4.1 Experiments on GT and Benchmarking with Contemporary Methods

The GT database comprises of scene-centric as well as non-reversible appearance centric conditions. This is a case of multi-variation. For instance, there is the presence of illumination, expression, scale, and pose, but the images belonging to a particular subject mostly vary by pose than by other factors. The PSP-plastic surgery procedure, BL-Blepharoplasty, SP-skin peeling, RY-Rhytidectomy, DE-Dermabrasion, OT-Otoplasty, BR-brow lift, RH-Rhinoplasty, OTO-others, CC-cheek&chin, EER-equal error rate, FAR-false acceptance rate experimental results are with respect to sLPP-LGE and LSDA-LGE and PCA plus LDA subspace learning methods (best performing subspace methods from previous analysis). From the identification and verification results given in Figure 4 and summarized in Table 4 the following observations are made.

**Table 1** Recognition performances of LEGGM descriptor and existing descriptors for different types of plastic surgery procedures

| PSP | Method | @ FAR 0.01 | @FAR 0.05 | @FAR 0.1 | EER | Rank-1 |
|---|---|---|---|---|---|---|
| BL | CLPB-M-S | 75.00 | 80.00 | 100 | 0.0921 | 65.00 |
| | CLBP-M | 60.00 | 70.00 | 80.00 | 0.1500 | 35.00 |
| | CLBP-S | 85.00 | 85.00 | 90.00 | 0.1000 | 55.00 |
| | LGBP | 73.29 | 90.10 | 95.05 | 0.0693 | 35.64 |
| | LEGGM | 71.29 | 86.14 | 89.11 | 0.1089 | 74.26 |
| SP | CLPB-M-S | 80.82 | 87.67 | 95.89 | 0.0818 | 31.51 |
| | CLBP-M | 60.27 | 80.82 | 82.19 | 0.1384 | 17.81 |
| | CLBP-S | 71.23 | 87.67 | 91.78 | 0.0947 | 23.23 |
| | LGBP | 94.52 | 97.26 | 100 | 0.0274 | 61.64 |
| | LEGGM | 83.33 | 88.89 | 93.06 | 0.0695 | 86.11 |
| RY | CLPB-M-S | 74.03 | 86.04 | 90.26 | 0.0973 | 15.26 |
| | CLBP-M | 52.60 | 69.81 | 76.62 | 0.1591 | 8.44 |
| | CLBP-S | 69.16 | 81.49 | 87.34 | 0.1135 | 18.48 |
| | LGBP | 84.42 | 93.83 | 95.78 | 0.0559 | 30.19 |
| | LEGGM | 76.62 | 87.99 | 92.53 | 0.0844 | 73.38 |
| DE | CLPB-M-S | 68.75 | 87.50 | 96.88 | 0.0670 | 28.13 |
| | CLBP-M | 53.13 | 75.00 | 84.38 | 0.1563 | 12.50 |
| | CLBP-S | 59.38 | 81.25 | 84.38 | 0.1250 | 34.38 |
| | LGBP | 87.50 | 90.63 | 90.63 | 0.0938 | 62.50 |
| | LEGGM | 75.00 | 84.38 | 87.50 | 0.1250 | 87.50 |
| OT | CLPB-M-S | 78.57 | 89.29 | 92.86 | 0.0893 | 30.36 |
| | CLBP-M | 67.86 | 80.36 | 82.14 | 0.1250 | 26.76 |
| | CLBP-S | 71.43 | 89.29 | 91.07 | 0.0899 | 26.76 |
| | LGBP | 78.57 | 87.50 | 89.29 | 0.1071 | 48.21 |
| | LEGGM | 76.36 | 89.09 | 90.91 | 0.0928 | 85.45 |
| BR | CLPB-M-S | 73.33 | 83.33 | 90.00 | 0.1000 | 28.33 |
| | CLBP-M | 60.00 | 75.00 | 75.00 | 0.1700 | 16.67 |
| | CLBP-S | 66.67 | 81.67 | 86.67 | 0.1169 | 28.33 |
| | LGBP | 85.00 | 91.67 | 93.33 | 0.0814 | 53.33 |
| | LEGGM | 58.33 | 78.33 | 83.33 | 0.1333 | 71.64 |
| RH | CLPB-M-S | 73.44 | 87.50 | 91.15 | 0.0889 | 20.31 |
| | CLBP-M | 68.23 | 82.29 | 88.02 | 0.1095 | 11.46 |
| | CLBP-S | 68.23 | 82.29 | 88.02 | 0.1095 | 11.46 |
| | LGBP | 83.33 | 94.27 | 95.31 | 0.0573 | 33.33 |
| | LEGGM | 78.65 | 88.54 | 92.71 | 0.0876 | 76.04 |
| OTO | CLPB-M-S | 81.69 | 91.55 | 95.77 | 0.0704 | 39.44 |
| | CLBP-M | 56.34 | 78.87 | 85.92 | 0.0985 | 25.35 |
| | CLBP-S | 69.01 | 85.92 | 90.14 | 0.1126 | 28.17 |
| | LGBP | 84.51 | 92.96 | 92.96 | 0.0704 | 60.56 |
| | LEGGM | 76.06 | 88.73 | 92.96 | 0.0739 | 83.10 |
| CC | CLPB-M-S | 76.19 | 95.21 | 100 | 0.0470 | 28.57 |
| | CLBP-M | 52.38 | 57.14 | 66.67 | 0.1905 | 19.05 |
| | CLBP-S | 71.43 | 85.71 | 90.48 | 0.0952 | 23.81 |
| | LGBP | 95.24 | 95.24 | 100 | 0.0476 | 71.43 |
| | LEGGM | 80.95 | 95.24 | 95.24 | 0.0476 | 80.95 |
| TOTAL | CLPB-M-S | 75.76 | 87.57 | 94.76 | 0.0815 | 31.88 |
| | CLBP-M | 58.98 | 73.37 | 80.10 | 0.1441 | 19.23 |
| | CLBP-S | 70.17 | 84.48 | 88.88 | 0.1064 | 26.62 |
| | LGBP | 85.15 | 92.61 | 94.90 | 0.0678 | 50.76 |
| | LEGGM | 75.18 | 87.48 | 90.82 | 0.1653 | 79.83 |

As expected of LSDA based on the preceding analysis, its identification rate at Rank-1 is far less in comparison to sLPP and PCA plus LDA. Its performance in this experiment may not be unconnected with the reason for its low performance in the preceding analysis, where the angle differences as a result of mirror-reflect images were observed to influence recognition with LSDA model. Here, the obvious cause for its performance is the presence of pose variation between images of a subject. It is common sense to say that the pose problem is similar to angle difference.

However, they are quite different in the sense that the angle difference image can assume the view of its mirror-reflect image (i.e., if a frontal view image the mirror-reflect images is still of frontal view but from an opposite direction), whereas with a change in pose the angle of interest in the images of a subject vary greatly. This explains why the Rank-1 identification rate of LSDA (in relation to the other subspace methods) in this experiment is far less than what it is in the previous analysis.

As can be further observed from Figure 4 and Table 4, PCA plus LDA performed far better than sLPP-LGE, which shows that the global structure information of data is very important in preserving appearance especially when pose is a major variation factor to contend with. Given the outcome of this experiment, a similar behavior of the subspace methods for LFW is expected.

The designed descriptor-based face recognition method in comparison with existing works in literature is observed from two perspectives, face alignment and classifier. The performance of the designed method in terms of identification rate is significantly higher than those of all other methods with no face alignment [1, 58, 59, 60]. It should also be noted that in these works more sophisticated classifiers such as, SVM [59], Bayes nearest neighbor classifier [61] and linear regression classifier (LRC) [58] where adopted. In [58] the original image without cropping is utilized, which means that the background information contributes to their recognition accuracy. For the method by Wouter *et al*. [29] with face alignment only a 0.80% difference is observed, though their result is obtained under different evaluation protocol that comprises of radial basis neural network classifier and only frontal-view images are used to train the subspace model. However, the proposed face recognition method adopted the NN, a baseline classifier, which in comparison with the method by Geng and Jiang [60] with a single feature classifier decision using the NN, a 6% increase in rank-1 identification rate is observed.

**Table 2** Recognition performance of LEGGM with various subspace learning models

| PSP | EM | ES-2 | | | | | EM | ES-1 | | | | |
|---|---|---|---|---|---|---|---|---|---|---|---|---|
| | | A | B | C | D | E | | A | B | C | D | E |
| BL | IR | 97.00 | 99.00 | 77.2 | 79.20 | 96.00 | IR | 89.10 | 89.10 | 91.10 | 85.10 | 89.10 |
| | VR | 99.01 | 99.01 | 92.07 | 90.11 | 99.26 | VR | 97.91 | 97.91 | 97.02 | 96.79 | 98.02 |
| | EER | 0.0099 | 0.0099 | 0.0793 | 0.0989 | 0.0074 | EER | 0.0209 | 0.0209 | 0.0298 | 0.0321 | 0.0198 |
| SP | IR | 98.60 | 98.60 | 90.30 | 94.40 | 98.60 | IR | 90.30 | 86.10 | 86.10 | 88.90 | 87.50 |
| | VR | 99.97 | 98.69 | 97.19 | 94.73 | 99.98 | VR | 95.83 | 94.43 | 94.43 | 94.44 | 95.84 |
| | EER | 0.0033 | 0.0130 | 0.0281 | 0.0527 | 0.0002 | EER | 0.0417 | 0.0557 | 0.0557 | 0.0556 | 0.0416 |
| RY | IR | 94.80 | 97.70 | 59.70 | 70.50 | 96.40 | IR | 80.80 | 73.40 | 73.10 | 76.30 | 84.10 |
| | VR | 97.50 | 99.35 | 84.74 | 84.73 | 99.32 | VR | 95.45 | 93.09 | 93.51 | 94.16 | 94.80 |
| | EER | 0.0260 | 0.0650 | 0.1526 | 0.1527 | 0.0680 | EER | 0.0455 | 0.0691 | 0.0649 | 0.0584 | 0.0520 |
| DE | IR | 100 | 100 | 100 | 100 | 100 | IR | 90.60 | 87.50 | 87.50 | 90.60 | 93.80 |
| | VR | 100 | 100 | 100 | 97.53 | 100 | VR | 96.93 | 96.87 | 96.87 | 93.75 | 94.20 |
| | EER | 0.0000 | 0.0000 | 0.0000 | 0.0247 | 0.000 | EER | 0.0307 | 0.0313 | 0.0313 | 0.0625 | 0.0580 |
| OT | IR | 100 | 100 | 90.90 | 96.40 | 96.40 | IR | 100 | 90.90 | 87.30 | 87.30 | 92.70 |
| | VR | 99.98 | 99.95 | 92.73 | 94.55 | 99.95 | VR | 96.36 | 94.55 | 96.35 | 96.13 | 94.78 |
| | EER | 0.0002 | 0.0005 | 0.0727 | 0.0545 | 0.0005 | EER | 0.0364 | 0.0545 | 0.0365 | 0.0387 | 0.0522 |
| BR | IR | 100 | 100 | 81.70 | 85.00 | 100 | IR | 85.00 | 81.70 | 81.70 | 83.30 | 80.00 |
| | VR | 98.47 | 98.33 | 90.00 | 86.67 | 99.97 | VR | 95.41 | 91.89 | 96.25 | 94.07 | 93.33 |
| | EER | 0.0153 | 0.0167 | 0.1000 | 0.1333 | 0.0003 | EER | 0.0459 | 0.0811 | 0.0375 | 0.0503 | 0.0667 |
| RH | IR | 97.40 | 97.90 | 75.00 | 82.20 | 97.40 | IR | 86.50 | 81.30 | 79.70 | 80.20 | 86.50 |
| | VR | 98.96 | 99.48 | 91.15 | 89.07 | 99.91 | VR | 96.90 | 93.34 | 94.27 | 94.76 | 95.32 |
| | EER | 0.0104 | 0.0520 | 0.0885 | 0.1093 | 0.0090 | EER | 0.0310 | 0.0660 | 0.0573 | 0.0524 | 0.0468 |
| OTO | IR | 98.60 | 100 | 90.1 | 87.30 | 98.60 | IR | 87.30 | 77.50 | 87.30 | 84.50 | 83.10 |
| | VR | 98.59 | 98.66 | 95.77 | 94.37 | 99.70 | VR | 95.77 | 96.90 | 95.77 | 95.76 | 96.83 |
| | EER | 0.0410 | 0.0134 | 0.0423 | 0.0563 | 0.0030 | EER | 0.0423 | 0.0310 | 0.0423 | 0.0424 | 0.0317 |
| CC | IR | 100 | 100 | 85.70 | 85.70 | 100 | IR | 85.70 | 85.70 | 85.70 | 85.70 | 85.70 |
| | VR | 100 | 100 | 95.71 | 96.07 | 100 | VR | 90.48 | 90.48 | 94.17 | 90.48 | 89.29 |
| | EER | 0.0000 | 0.0000 | 0.0429 | 0.0393 | 0.0000 | EER | 0.0952 | 0.0952 | 0.0583 | 0.0952 | 0.1071 |
| TOTAL | IR | 98.49 | 99.24 | 83.40 | 86.74 | 98.16 | IR | 88.37 | 83.69 | 74.87 | 84.66 | 86.94 |
| | VR | 99.16 | 99.27 | 93.26 | 91.98 | 91.98 | VR | 95.68 | 94.38 | 95.40 | 94.48 | 94.71 |
| | EER | 0.0118 | 0.0189 | 0.0674 | 0.0802 | 0.0802 | EER | 0.0433 | 0.0561 | 0.0460 | 0.0542 | 0.0529 |

### 5.4.2 Experiments on LFW database and Benchmarking with Contemporary Methods

First of all, let us recall that the data set that forms this database is of gross variation in pose, illumination, noise, blur, expression, occlusion and lots more but most importantly it varies by pose.

It is also important to note that the data set is a subset of the LFW data set which originally is experimented on for verification benchmarking. Here, a deviation from the known benchmarking protocol is made, but in relation to these literatures [63-64]. Since only these literatures have used this LFW identification protocol, the result of the designed descriptor-based face recognition method is compared with the results of these literatures and some others that utilized it for verification task. The results of the designed descriptor-based face recognition method are obtained in this work, while the results of the methods of other researchers are duly cited from their respective work.

Here, the results of experiments with ES-1 and ES-2 evaluation scenarios described earlier in this section are plotted in the same graph. The results are provided in Figure 5 and tabulated in Table 5. From Figure 5, two diverse results can be observed between ES-1 and ES-2, one can quickly say it is because of the number of images used in the two experiments. However, a more theoretical reason is that in ES-1 the system is provided with more samples which enable the system to achieve good generalization for unseen test data. Recall that in both ES-1 and ES-2, the same number of images make-up the gallery and probe (test) set. However, their results differ because in ES-2, only fewer samples were used to train the system and so the system could not generalize properly to the unseen test image.

Now, from the perspective of the linear subspace learning models, it can be seen that the same observations made in the previous experiments still hold here. The only difference is that in this experiment LSDA failed because a Rank-1 identification rate of 5.79% (ES-2) and 11.11% (ES-1) clearly indicates that it almost did not identify any individual. The only consolation is that from Table 5, its verification capability is shown to be above average even though at various FAR (0.01, 0.05 and 0.1) it still performed poorly.

Since the data set that forms the LFW database images varies grossly by pose, it is of expectation that in both scenarios of ES-1 and ES-2 the performances of PCA plus LDA subspace learning from LEGGM will perform better than sLPP. However, the observation made in this experiment shows it only stands true for ES-1 than ES-2 because the images of a subject used for ES-1 rarely have frontal-view images. The point that is intended to be made here is that sLPP provides better fit to LEGGM data only when the face images of a subject follow a particular trend, that is, the pose of the images are all either non-frontal or frontal and not both.

With the inclusion of the results of contemporary methods obtained on the LFW158 and LFW610 in Table 5, it can be seen that the designed descriptor-based face recognition method achieves better or similar results with reasons. The performance of the designed descriptor-based method in terms of verification

**Table 3** Performance comparisons of the descriptor-based face recognition method with existing methods for plastic surgery altered faces

| | Method | Rank-1 (%) | Comment | VR (%) | EER (%) |
|---|---|---|---|---|---|
| Designed | LEGGM-sLPP-LGE | 88.70 | -single classifier decision<br>-tested on the entire database<br>-no face alignment | 96.33 | 3.68 |
| | LEGGM-sLPP-OLGE | 83.44 | -single classifier decision<br>-tested on the entire database<br>-no face alignment | 94.87 | 3.68 |
| | LEGGM-LSDA-LGE | 84.23 | -single classifier decision<br>-tested on the entire database<br>-no face alignment | 95.56 | 4.44 |
| | LEGGM-LSDA-OLGE | 85.76 | -single classifier decision<br>-tested on the entire database<br>-no face alignment | 94.98 | 4.91 |
| | LEGGM-PCA+LDA | 87.10 | -single classifier decision<br>-tested on the entire database<br>-no face alignment | 95.39 | 4.61 |
| Others | Near-Set [50] | 55.55 | -geometric-based approach | n/a | n/a |
| | FACE [51] | 70.00 | tested on the entire database | n/a | 24.00 |
| | FACE [52] | 85.40 | tested on the entire database | n/a | 7.20 |
| | Sparse-based [53] | 77.90 | -classifier score multi-feature fusion<br>-score fusion from multi-components | n/a | n/a |
| | Granular-based [54] | 78.61 | -tested on 60% of the database<br>-score fusion from multi-feature<br>-face alignment | n/a | n/a |
| | Granular-based [55] | 87.32 | -tested on 60% of the database<br>-score fusion from multi-feature<br>-face alignment | n/a | n/a |
| | GPROF [30] | 86.11 | -tested on the entire database<br>-face alignment<br>-score fusion from multi-feature | 68.69 | 31.32 |
| | Multimodal [47] | 70.30 | -tested on the entire database | | |
| | | 73.90 | 661 subjects | | |
| | | 87.40 | classifier score multi-feature fusion<br>661 subjects | n/a | n/a |
| | SSIM [49] | 77.55 | -face alignment<br>-784 subjects/classifier fusion<br>-classifier score fusion | n/a | n/a |
| | GFRPS [56] | 77.30 | -geometric based approach | n/a | n/a |
| | Region-based [57] | 67.55 | 814 subjects | | |

EER-equal error rate, VR-verification rate, n/a-not applicable

**Table 4** Recognition performance of the descriptor-based face recognition method with contemporary methods for the case of GT database

| | Method | Rank-1 (%) | @FAR 0.01 | @FAR 0.05 | @FAR 0.1 | EER (%) | VR (%) |
|---|---|---|---|---|---|---|---|
| Designed | LEGGM-sLPP-LGE | 96.00 | 94.00 | 98.00 | 98.00 | 3.90 | 96.10 |
| | LEGGM-LSDA-LGE | 78.00 | 70.00 | 82.00 | 88.00 | 10.04 | 89.96 |
| | LEGGM-PCA plus LDA | 98.00 | 98.00 | 98.00 | 98.00 | 1.09 | 98.91 |
| Others | Majumdar et al. [59] | 86.5 | n/a | n/a | n/a | n/a | n/a |
| | Maturana et al. [61] | 92.57 | n/a | n/a | n/a | n/a | n/a |
| | Naseem et al. [58] | 92.86 | n/a | n/a | n/a | n/a | n/a |
| | Geng and Jiang [60] | 97.43 | n/a | n/a | n/a | n/a | n/a |
| | Li et al. [62] | 96.9 | n/a | n/a | n/a | n/a | n/a |
| | Wouter et al. [26] | 98.80 | n/a | n/a | n/a | n/a | n/a |

FAR-false acceptance rate, EER-equal error rate, VR-verification rate, n/a-not applicable rates is substantially higher than those of all other methods in both scenarios, ES-1 and ES-2, despite being obtained with a weighted NN classifier. It is generally known that the more the sophistication of a classifier the better the classification performance and that a good feature can make the least classifier to be effective [65]. For instance, consider the designed descriptor-based method and the generalized region assigned binary (GRAB) in ES-2, it can be seen that the designed method outperforms GRAB (NN) [64] by 15.74%. However, upon the use of GRAB (SVM) [64] a performance increase by 21.64% is observed for GRAB. In ES-1, aside Shen et al. [66], the designed method came out top-best performer both in verification and identification experiments. Shen et al. [67] is reported alongside Shen et al. [66] because they are of the same classifier, but in [66], further classifier decision is from dictionary of features. The dictionary of features is known to improve classification accuracy [68].

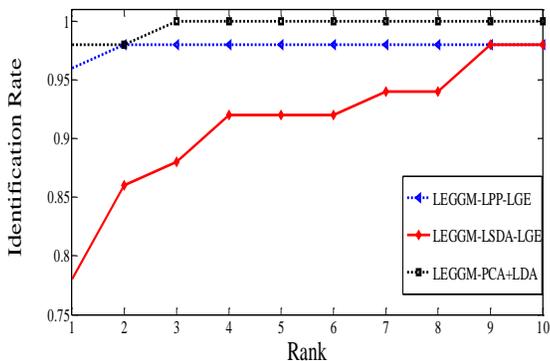

**Figure 4** Identification performance of the descriptor-based face recognition method for GT database

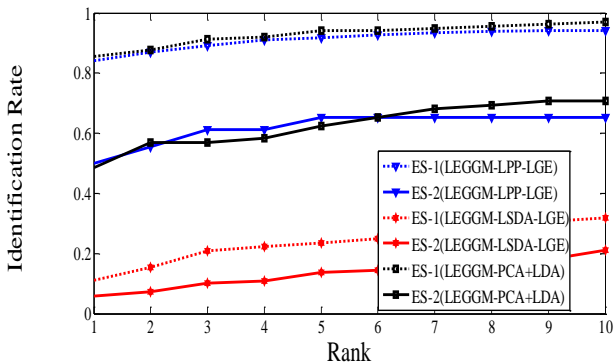

**Figure 5** Identification performance of the descriptor-based face recognition method for the LFW database

### 5.4.3 Experiments on Heterogeneous Database

This database is formed from the heterogeneous data set, which is borne out of the need to formulate a real-world scenario where the face recognition system is unaware of any faces that have undergone plastic surgery procedures. Therefore, the surgery images and non-surgery images of different subjects are combined. Here, it is of expectation that the designed descriptor-based face recognition method will be robust against a number of image formation factors that are present in the system. Like in all the experiments, the results of designed descriptor-based face recognition methods are on the basis of the subspace learning methods. The results are reported in terms of identification rate, verification rate and EER. The plots of the results are shown in Figure 6 and Table 6.

Again, it is expected that the performance of subspace learning from LEGGM using LSDA will be below the performances of PCA plus LDA and sLPP. This is for the fact that the percentages of the images collected from GT and LFW From Figure 6 and Table 6, it can be seen that the use of PCA plus LDA performed best in all the experiments by a large margin, which can be observed from the Rank-1 up to Rank-10. The use of sLPP performed second best followed by LSDA. In comparison with the previously reported experiments, LSDA can be seen to have significant increase in recognition accuracy. The obvious reason one could point at is the fact that there are more percentages of frontal-view images in the heterogeneous database than is included in the other databases (GT or LFW).

data sets are of view-point (pose) differences. However, it is the extent to which the performance of one subspace learning model from LEGGM deviates from the other that will be the object of our discussion. That notwithstanding, far better recognition accuracies are envisaged to be achieved for the entire system if the image sets in the database are restricted to only the frontal-view.

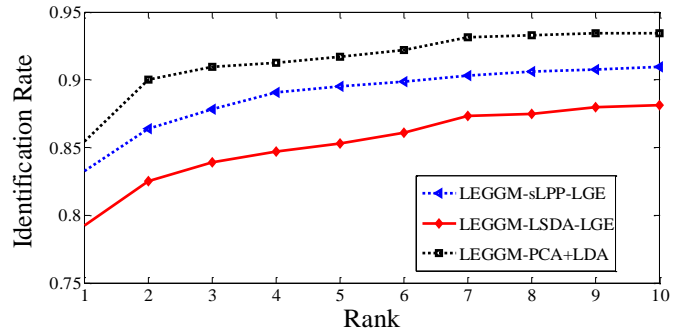

**Figure 6** Identification performance of the descriptor-based face recognition method for the heterogeneous database

Overall, the experiment on the heterogeneous data sets validates that the intrinsic facial characteristics of the descriptor-based face recognition method captured and retains for recognition can, to a good extent, be robust against real-world face recognition variation factors.

### 5.5 *Discussion*

The need to design and develop the new LEGGM descriptor is not so that it replaces the Gabor descriptor, but to overcome its limitations. Our arguments can be observed from experimental point of view.

The performance analysis of LEGGM, being a variant of the Gabor descriptor, will not be complete without comparing it with Gabor itself. Presented here in Figure 7 is the experimental result of Gabor (in its originality) and LEGGM (a variant of Gabor). The results show the increase recognition performance of LEGGM descriptor over the well-known Gabor descriptor. From the CMC graph of Figure 7, it can be seen that LEGGM outperformed the Rank-1 identification rate of Gabor for the GT database by 6%, while 11.19% for the plastic surgery database. Therefore, it can be emphatically stated that LEGGM is highly effective for describing face patterns than the Gabor though this claim is based on the presented data to the descriptor based face recognition system. Further works may conduct analysis on various levels of face formation factors and degree of variation in order to validate our claim.

### 5. Conclusion

The experimental results reported in this paper point to a fact that the success of face recognition system is highly dependent on the face representation approach. The success of LEGGM descriptor is attributed to the PISP defined global appearance information. Through experiments we were able to see that even a simple classifier can emerge powerful enough where patterns have been well-defined.

**Table 5** Recognition performance of the descriptor-based face recognition method with Contemporary Methods for the case of LFW database

|  |  | ES1 | | | | | |
|---|---|---|---|---|---|---|---|
|  | Method | @FAR 0.01 | @FAR 0.05 | @FAR 0.1 | EER | VR | Rank-1 |
| LFW158 Designed | LEGGM-sLPP-LGE | 84.06 | 92.39 | 94.93 | 6.17 | 93.83 | 84.06 |
|  | LEGGM-LSDA-LGE | 02.17 | 10.14 | 18.12 | 41.89 | 58.11 | 11.11 |
|  | LEGGM-PCA+LDA | 90.58 | 97.10 | 97.83 | 3.62 | 96.38 | 85.51 |
| Others | Fernandez and Vicente [69] | 36.11 | 52.78 | 62.76 | 20.48 | 79.52 | n/a |
|  | Liang et al. [70] | 60.00 | 72.61 | 85.25 | 11.68 | 88.32 | n/a |
|  | Shen et al.. [66, 67] | n/a | n/a | n/a | n/a | n/a | 80.70/88.30 |

|  |  | ES-2 | | | | | |
|---|---|---|---|---|---|---|---|
|  | Method | @FAR 0.01 | @FAR 0.05 | @FAR 0.1 | EER | VR | Rank-1 |
| LFW610 Designed | LEGGM-sLPP-LGE | **51.39** | 61.11 | 66.67 | 25.01 | 74.99 | **50.00** |
|  | LEGGM-LSDA-LGE | 08.33 | 28.61 | 33.33 | 38.78 | 61.22 | 05.79 |
|  | LEGGM-PCA+LDA | 50.00 | **62.50** | 72.22 | 20.81 | 79.19 | 48.61 |
| Others | GRAB (k-NN) [64] | n/a | n/a | n/a | n/a | n/a | 34.26 |
|  | GRAB (SVM) [64] | n/a | n/a | n/a | n/a | n/a | 55.90 |
|  | V1-like [64] | n/a | n/a | n/a | n/a | n/a | 41.30 |

ES-1-evaluation scenario-1&2 (this protocol is a replica of the work in [63,64]), n/a-not applicable

**Table 6** Performance of the descriptor-based face recognition method in a heterogeneous case

| Method | @FAR 0.01 (%) | @FAR 0.05 (%) | @FAR 0.1 (%) | EER (%) | VR (%) | Rank-1 (%) |
|---|---|---|---|---|---|---|
| LEGGM-sLPP-LGE | 88.12 | 92.34 | 94.69 | 6.86 | 93.14 | 83.28 |
| LEGGM-LSDA-LGE | 83.13 | 89.06 | 91.56 | 8.89 | 91.11 | 79.22 |
| LEGGM-PCA+LDA | **93.44** | **95.78** | **96.88** | **4.21** | **96.79** | **85.47** |

FAR-false acceptance rate, EER-equal error rate, VR-verification rate

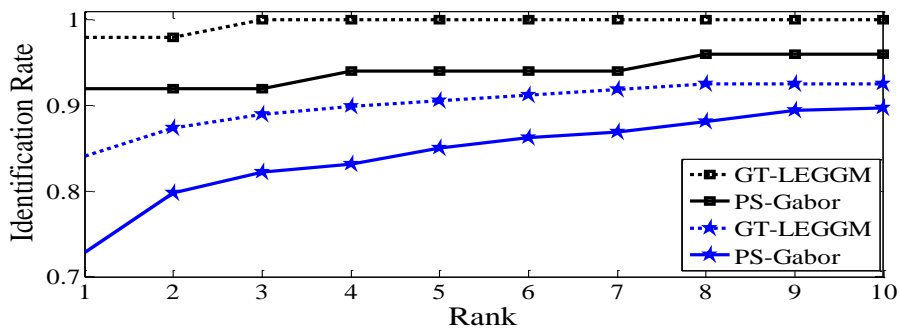

**Figure 7** Face recognition performances on plastic surgery database and GT database using the Gabor facial descriptor and LEGGM facial descriptor

The extensive experimental analysis and comparison of the proposed descriptor-based face recognition method with the contemporary methods show the following: On the plastic surgery data set, unlike the contemporary methods that merged classification decision, we adopted a single classifier decision using the non-parametric nearest neighbor classifier and was able to achieve best performance recorded so far. On the GT and LFW data sets, competitive recognition rates were achieved. On the heterogeneous data set, the designed descriptor-based face recognition method showed to be insensitive to the image formation factors. In the future work, we will like to investigate on the family of the wavelets because it has been found that the wavelets family decompose a signal into non-overlapping sequences of information that easily express sharp edges, lines and shape details of a signal. However, Gabor is deficient in such regards. Therefore, it will be interesting to analyze and investigate how other families of wavelets can interpret $\chi$ at various scales and frequencies.

**Appendix A.** Technical detail of frequency characteristics of Complete Face Structural Pattern over Gray-level information

The FFT of the resulting image of (a) shows a well distributed frequency spectrum information (see Figure 8 (a1-a2)), which signifies that there are more structured high-level (edge) information in the image as opposed to the grey-level information (see Figure 8 (b1-b2)), which shows a concentrate at the center and mostly capture low level information (texture).

**Appendix B.** List of Abbreviations for Subspace learning from LEGGM

The following abbreviations are made,
A-ES-2: LEGGM-LPP-LGE
A-ES-1: LEGGM-LPP-LGE
B-ES-2: LEGGM-LPP-OLGE
B-ES-1: LEGGM-LPP-OLGE
C-ES-2: LEGGM-LSDA-LGE
C-ES-1: LEGGM-LSDA-LGE
D-ES-2: LEGGM-LSDA-OLGE
D-ES-1: LEGGM-LSDA-OLGE
E-ES-2: LEGGM-PCA+LDA
E-ES-1: LEGGM-PCA+LDA.

The following Figure 2 are the figure illustrations of the tabulated accuracies in Table 2.

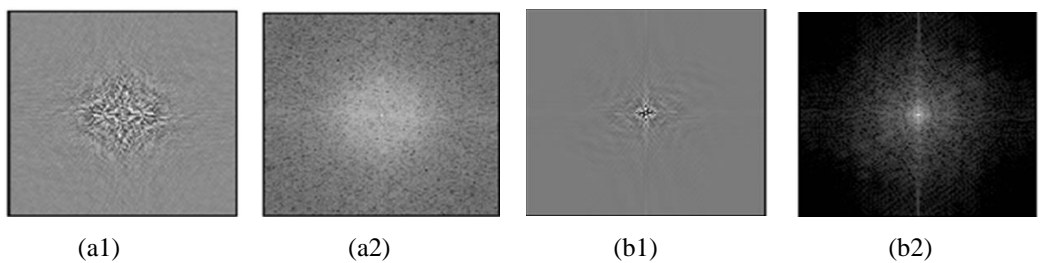

(a1)　　　(a2)　　　(b1)　　　(b2)

**Figure 8 Fourier** spectrum analysis of face signals. The spectrum of the complete face structural image and the spectrum of the grey-level image (a1 and b1), respectively. The complex FFT of the signals (a2 and b2), respectively.

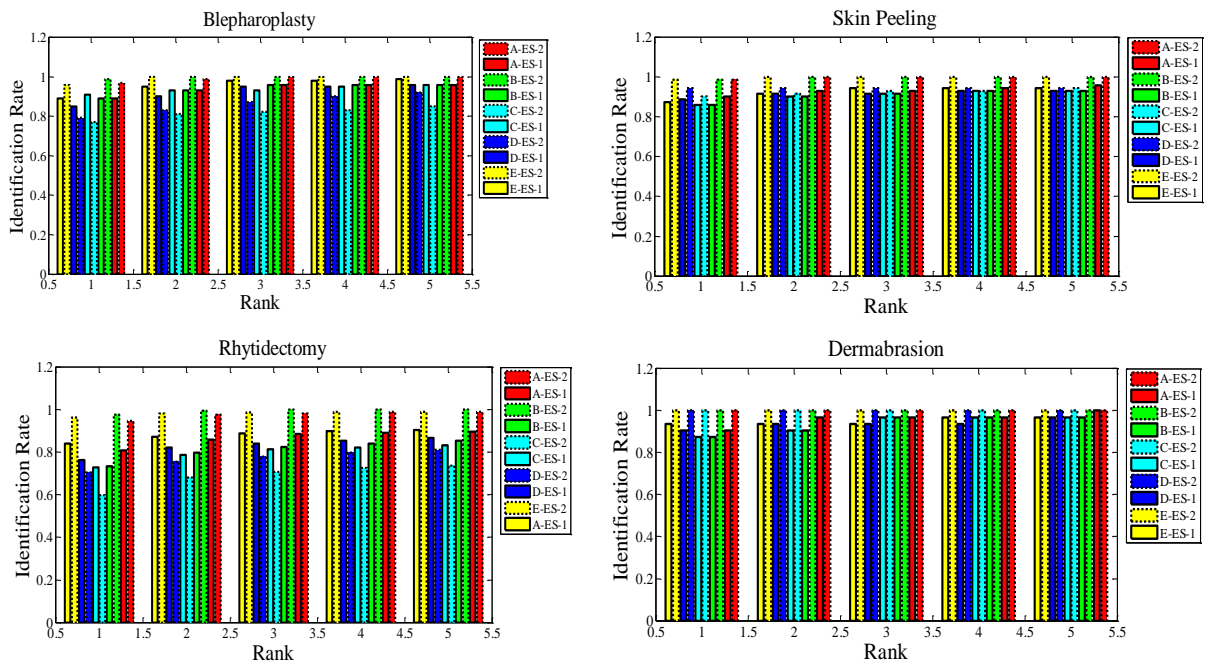

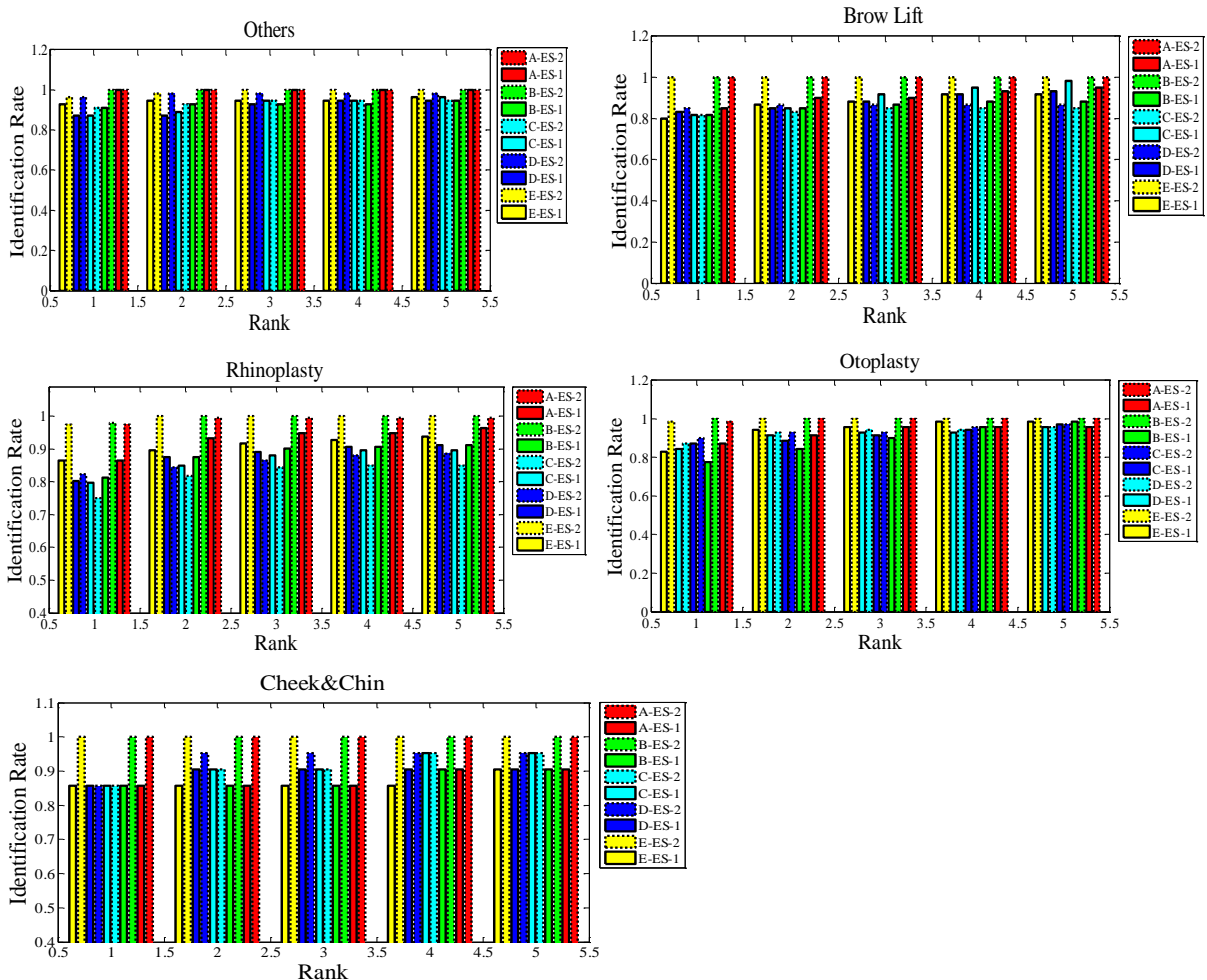

**Figure 9** Identification performance of LEGGM with subspace learning for different plastic surgery procedures